# Hybridizing PSM and RSM Operator for Solving NP-Complete Problems: Application to Travelling Salesman Problem

Otman ABDOUN, Chakir TAJANI and Jaafar ABOUCHABKA
LaRIT, Department of Computer Science
IBN Tofail University, Kenitra, Morocco
Otman.fsk@gmail.com, Chakir_tajani@hotmail.fr, Aboucha06-univ@yahoo.fr

**Abstract**

In this paper, we present a new mutation operator, Hybrid Mutation (HPRM), for a genetic algorithm that generates high quality solutions to the Traveling Salesman Problem (TSP). The Hybrid Mutation operator constructs an offspring from a pair of parents by hybridizing two mutation operators, PSM and RSM. The efficiency of the HPRM is compared as against some existing mutation operators; namely, Reverse Sequence Mutation (RSM) and Partial Shuffle Mutation (PSM) for BERLIN52 as instance of TSPLIB. Experimental results show that the new mutation operator is better than the RSM and PSM.

*Keywords:* NP-complete problem, Traveling Salesman Problem, Mutation operator.

## 1. Introduction

NP-Complete is a class of problems that are so difficult that even the best solutions cannot consistently determine their solutions in an efficient way. Specifically, NP Complete problems can only possibly be solved in polynomial time using a nondeterministic Turing machine. Problems such as the Traveling Salesman problem fall into the realm of NP Complete problems. Because of the problem with finding the "best" solution, programs are often developed to find a usually reasonable solution.

The traveling salesman problem (TSP) is a well known and important combinatorial optimization problem. In the traveling salesman problem (TSP) we are given *n* vertices *1, .., n* and all *n(n-1)/2* distances between them, as well as a budget *b*. We are asked to find a tour, a cycle that passes through every vertex exactly once, of total cost b or less or to report that no such tour exists. The number of solutions becomes extremely large for even moderately large n so that an exhaustive search is impracticable.

The methods that provide the exact optimal solution to the problem are called exact methods. An implicit way of solving the TSP is simply to list all the feasible solutions, evaluate their objective function values and pick out the best. However it is obvious that this "exhaustive search" is grossly inefficient and impracticable because of vast number of possible solutions to the TSP even for problem of moderate size. Since practical applications require solving larger problems, hence emphasis has shifted from the aim of finding exactly optimal solutions to TSP, to the aim of getting, heuristically, 'good solutions' in reasonable time and 'establishing the degree of goodness'. Genetic algorithm (GA) is one of the best heuristic algorithms that have been used widely to solve the TSP instances [1].

In Genetic Algorithm, a population of potential solutions termed as chromosomes and individuals are evolved over successive generations using a set of genetic operators called selection, crossover and mutation. First of all, based on some criteria, every chromosome is assigned a fitness value and then the selection operator is applied to choose relatively fit chromosomes to be part of the reproduction process. In reproduction process new individuals are created through application of operators. Large number of operators has been developed for improving the performance of GA, because the performance of algorithm depends on the ability of these operators [15]. One of the operators, mutation operator is used to maintain adequate diversity in the population of chromosomes and avoid premature convergence. Whereas crossover operator, blends the genetic information between chromosomes to explore the search space[16].

Researches have shown that the mutation operator plays an important role in genetic algorithm, so many mutation operators have been proposed for the TSP. Partial Shuffle Mutation (PSM) [4], Inversion Mutation [8], Exchange mutation [9], Displacement Mutation [10], Insertion Mutation [11], Heuristic mutation [12], Greedy Swap Mutation (GSM) [13], Reverse Sequence Mutation (RSM), etc.

In this paper, a new mutation operator named Hybrid Mutation (HPRM) is developed for solving the TSP.

This paper is organized as follows: Section 2 presents the NP-complete problem. Section 3 proofs TSP is an NP-complete problem. Section 4 presents the proposed operator. Section 5 describes computational experiments for four mutation operators. Finally, Section 6 presents comments and concluding remarks.

## 2. NP-Complete Problem

NP-complete problems are a class of hard problems, which so far cannot be solved in polynomial time. Many problems from our everyday life are NP-complete, and although we might not be able to find an optimal solution within reasonable time, different methods exist to find a satisfactory solution. These methods include approximation, where a near optimal solution can be guaranteed, randomization, where an optimal solution can be found with a certain probability, or heuristics where a good solution can be found, but where there is no guarantee, it will be found fast. For many problems it is difficult to devise a complete algorithm that guaranties an optimal solution. A higher level of heuristics called meta-heuristics can then be used to combine the solution given by heuristics and solution strategies to obtain a better solution.

Problems that can be solved by algorithms in polynomial time are considered to be so called easy problems. For a problem of size *n* the time needed to find a solution is a polynomial function of *n*. Harder problems requires on the other hand an exponential function of *n*, which of course means that the execution time grows much faster than for an easy problem, when the size of the problem increases. NP-complete problems are hard problems to solve. They belong to a class of computational problems, for which no deterministic polynomial algorithm has been found.

NP-complete problems are a subset of the class *NP* (Non-Deterministic Polynomial). A Non-deterministic algorithm is able to find a correct solution, but it is not always guaranteed. The solution is found by making a series of guesses, and the algorithm will only arrive at a correct solution, if the right guesses are made along the way. A problem is called *NP*, if its solution can be found and verified by a non-deterministic algorithm in polynomial time. The class has the following definition according to [5]:

*__Definition:__* A yes-no-problem is in *NP* if there is a polynomial *p* and a randomized.
*p*-bounded algorithm *A* such that for every input *X* the following holds:
True answer for *X* is **YES** then PR[A(X;R) = YES] > 0
True answer for *X* is **NO**  then PR[A(X;R) = YES] = 0
where P*R*[Z] denotes the probability of event *Z* over uniform distribution of *R*.

The list of *NP*-complete is long, there exits several thousand problems, such as The Travelling Salesman problem (TSP), Constraint Satisfaction Problems (CSP), A Satisfiability (SAT), The Job Shop Scheduling problem, The k-Graph Partitioning problem, The vertex cover problem, etc.

The question now is, the Travelling Salesman Problem is an NP-Complete problem?

## 3. TSP is an NP-complete problem?

The TSP is a classical NP-Complete combinatorial optimization problem which comes up in different situations in our world. The TSP was first introduced by Karl Menger, in Vienna, and Harvard Universities and its significance was raised at Princeton University in 1930's. [2]. It is also the most studied problem for finding optimal solution. It can be stated as – *N* points (cities) as well as the cost of traveling between each pair of them are given. Assume that a sales person starting from a given city has to visit each city exactly once and should come back to the starting city to complete the tour. The aim is to find out the optimum tour in which the total cost is minimized. More formally TSP can be formulated as problem of graph theory. Given a graph *G* on a set of *N* vertices and the distances (costs of travelling) between each pair of cities as cost matrix matrix *C = [cij]* of order *N* associated with ordered node pairs *(i, j)*. The objective is to find a close tour with a minimum-cost that visits each city once returning to the starting city i.e. finding the shortest Hamiltonian cycle through *G*. On the basis of the structure of the cost matrix, the TSPs are classified into two groups symmetric and asymmetric. The TSP is symmetric if *Cij = Cji*, for all *i, j* and asymmetric otherwise. For an n-city asymmetric TSP, there are *(N−1)!* Possible solutions, one or more of which gives the minimum cost. For an n-city symmetric TSP, there are *(N −1)!/2* possible solutions along with their reverse cyclic permutations having the same total cost. In either case the number of solutions becomes extremely large for even moderately large *N* so that an exhaustive search is impracticable. Since TSP is *NP* complete the corresponding optimization problems are therefore NP hard.

The travelling salesman problem (TSP) is an NP-hard problem in combinatorial optimization studied in operations research and theoretical computer science [5].

*__Theorem:__* The subset-sum problem is NP-complete [6].

*__Proof :__* We first show that TSP belongs to NP. Given an instance of the problem, we use as a certificate the sequence of n vertices in the tour. The verification algorithm checks that this sequence contains each vertex exactly once, sums up the edge costs, and checks whether the sum is at most k. This process can certainly be done in polynomial time.

To prove that TSP is NP-hard, we show that *__HAM-CYCLE ≤ P__* TSP. Let *G = (V, E)* be an instance of HAM-CYCLE. We construct an instance of TSP as follows. We form the complete graph *G' = (V, E'),* , where *E'={ (i,j) :  i, j ∈ V  and  i ≠j }*, and we define the cost function *c* by

$$c(i,j) = \begin{cases} 0 & if \ (i,j) \in E \\ 1 & if \ (i,j) \notin E \end{cases} \qquad (5)$$

(Note that because *G* is undirected, it has no self-loops, and so *c(v, v)=1* for all vertices *v*∈*V*.) The instance of TSP is then *(G', c, 0),* which we can easily create in polynomial time.

We now show that graph *G* has a Hamiltonian cycle if and only if graph *G'* has a tour of cost at most *0*. Suppose that graph *G* has a Hamiltonian cycle *h*. Each edge in *h* belongs to *E* and thus has cost *0* in *G'*. Thus, *h* is a tour in *G'* with cost *0*.

Conversely, suppose that graph *G'* has a tour *h'* of cost at most *0*. Since the costs of the edges in *E'* are *0* and *1*, the cost of tour *h'* is exactly *0* and each edge on the tour must have cost *0*. Therefore, *h'* contains only edges in *E*. We conclude that *h'* is a Hamiltonian cycle in graph *G*.

The best known algorithms have exponential (deterministic) rum time complexity. Such combinatorial optimization problems are in the domain of Genetic algorithms that's why TSP has been solved through GA though, as far as the heuristic approach is concerned, it has been provided many algorithms for finding near optimal solutions for symmetric as well as asymmetric TSP.

Genetic algorithms (GAs) [17] are population based search techniques which mimics the principles of natural selection and natural genetics laid by Charles Darwin. Since Holland introduced the genetic algorithm (GA) in the early 1970's, many researchers have become interested in it as a new method of solving real life problems. As it is a promising heuristic approach to locate near optimal solution in large spaces, it is not surprising that it is a target of many researchers [16].

In Genetic Algorithm, a population of potential solutions termed as chromosomes and individuals are evolved over successive generations using a set of genetic operators called selection, crossover and mutation. Large number of operators has been developed for improving the performance of GA, because the performance of algorithm depends on the ability of these operators [15]. One of the operators, mutation plays the role of recovering the lost genetic materials as well as for randomly disturbing genetic information and to maintain adequate diversity in the population of chromosomes [16].

## 4. Hybrid Mutation Operator

Mutation involves the modification of the value of each 'gene' of a solution with some probability pm, (the mutation probability). The role of mutation in GAS has been that of restoring lost or unexplored genetic material into the population to prevent the premature convergence of the GA to suboptimal solutions.

Mutation plays the role of recovering the lost genetic materials as well as for randomly disturbing genetic information. Mutation prevents the algorithm to be trapped in a local minimum. It is an insurance policy against the irreversible loss of genetic material. Mutation is viewed as a background operator to maintain genetic diversity in the population. It introduces new genetic structures in the population by randomly modifying some of its building blocks. Mutation helps escape from local minima's trap and maintains diversity in the population.

There are many different forms of mutation for the different kinds of representation. For binary representation, a simple mutation can consist in inverting the value of each gene with a small probability. The probability is usually taken about 1/L, where L is the length of the chromosome. It is also possible to implement kind of hill-climbing mutation operators that do mutation only if it improves the quality of the solution. Such an operator can accelerate the search. But care should be taken, because it might also reduce the diversity in the population and makes the algorithm converge toward some local optima.

Partial Shuffle Mutation (PSM) [4] changes part of the order of the genes in the genotype. Inversion Mutation [8] selects two positions within a chromosome/tour and then inverts the substring between these two positions. Exchange mutation by [9] selects two positions at random and swaps the cities on these positions. Displacement Mutation [10] takes a sub tour at random and inserts it in a random. Insertion Mutation [11] selects a city at random and inserts it in a random position. Insertion can be viewed as a special case of displacement in which the substring contains only one city. Heuristic mutation by [12] was designed with neighborhood technique in order to produce an improved offspring. Greedy Swap Mutation (GSM) [13] also selects better result and therefore comes up closer to the solution quickly and Reverse Sequence Mutation (RSM).

Here two most known existing mutation operators are explained below. In addition to these the proposed mutation operators will be presented. The two most known existing mutation operators are Partial Shuffle Mutation (PSM) and Reverse Sequence Mutation (RSM) [4]. The proposed mutation operator is Hybridizing PSM and RSM Mutation operator (HPRM).

### 4.1 The most known existing mutation operators

a- Reverse Sequence Mutation (RSM)

In the reverse sequence mutation operator, we take a sequence S limited by two positions *i* and *j* randomly chosen, such that *i<j*. The gene order in this sequence will be reversed by the same way as what has been covered in the previous operation. The algorithm (Fig. 5) shows the implementation of this mutation operator.

Table 1: Mutation operator RSM

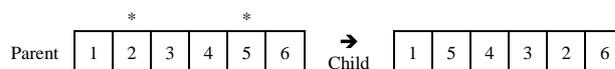

```
Input: Parents   x₁=[x_{1,1},x_{1,2},……,x_{1,n}]
Output: Children  x₁=[x_{1,1},x_{1,2},……,x_{1,n}]
-------------------------------------------------------
Choose two mutation points a and b such that 1 ≤ a ≤ b ≤ n;
Repeat
          Permute (x_a, x_b);
          a = a + 1;
          b = b − 1;
until  a<b
```

Fig. 1. Algorithm of RSM operator

b- Partial Shuffle Mutation (PSM)

The Partial Transfer Shuffle (PSM) as its name suggests, change part of the order of the genes in the genotype. The algorithm (Fig. 6) describes in detail the stages of change.

```
Input: Parents x=[x₁,x₂,……,xₙ]
       and Pm is Mutation probability
Output: Children x=[x₁,x₂,……,xₙ]
-------------------------------------------------------
i = 1;
Repeat
    Choose p a random number between 0 and 1
    if  p < Pm  then
      Choose j a random number between 1 and n;
          Permute (x_i, x_j);
      End if
Until i ≤ n
```

Fig. 2. Algorithm of Mutation operator PSM

4.2  Hybridizing PSM and RSM Operator (HPRM)

The Hybridizing PSM and RSM Operator (HPRM) constructs an offspring from a pair of parents by hybridizing two mutation operators, PSM and RSM. The algorithm (Fig. 3) presents the stages of change in our proposed operator.

```
Input: Parents x=[x₁,x₂,……,xₙ] and Pm is Mutation probability
Output: Children x=[x₁,x₂,……,xₙ]
-------------------------------------------------------
Choose two mutation points a and b such that 1 ≤ a ≤ b ≤ n;
Repeat
            Permute (x_a, x_b);
    Choose p a random number between 0 and 1
      if  p < Pm  then
      Choose j a random number between 1 and n;
          Permute (x_a, x_j);
      End if
          a = a + 1;
          b = b − 1;
Until a < b
```

Fig. 3. Algorithm of the proposed mutation operator, HPRM

## 5. Result and discussion

In this section the results of the proposed operator HPRM and the best mutation operators, RSM and PSM, are compared and discussed. To resolve a real Travelling Salesman Problem, we use the test problem BERLIN52 to 52 locations in the city of Berlin (Fig. 4). The only optimization criterion is the distance to complete the journey. The optimal solution to this problem is known, it's 7542 m (Fig. 5).

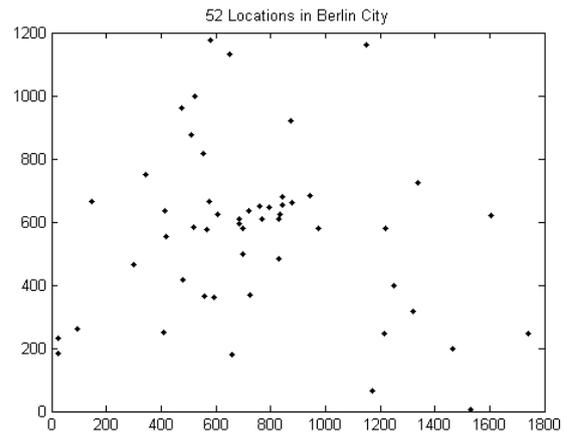

Fig. 4. The 52 locations in the Berlin city

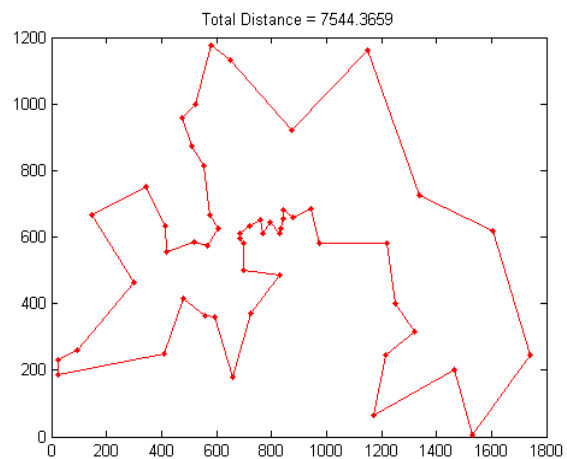

Fig. 5. The optimal solution of Berlin52

We change at a time one parameter and we set the others and we execute the genetic algorithm fifty times. The programming was done in *C++* on a PC machine with *Core2Quad 2.4GHz* in **CPU** and *2GB* in **RAM** with a *CentOS 5.5 Linux* as an **operating system**.

To compare statistically the three operators, HPRM, RSM and PSM, these are tested one by one on 50 different initial populations after that those populations are reused for each operator. In the case of the comparison of crossover operators, the evolutionary algorithm is presented in Figure 6 which the operator of variation is given by the crossover algorithm OX [18] and the selection is made by Roulette for choosing the shortest route.

```
Generate the initial population P₀
i = 0
Repeat
    P'ᵢ = Variation (Pᵢ);
    Evaluate (P'ᵢ);
    Pᵢ₊₁ = Selection ([P'ᵢ, Pᵢ]);
Until i<Itr
```

Fig.6. Evolutionary algorithm

Figure 7 shows the statistics of the experiments relating to the operators of mutation. The obtained results indicate the performance of the HPRM operator to found the better minimal values than the remaining mutations used in this work.

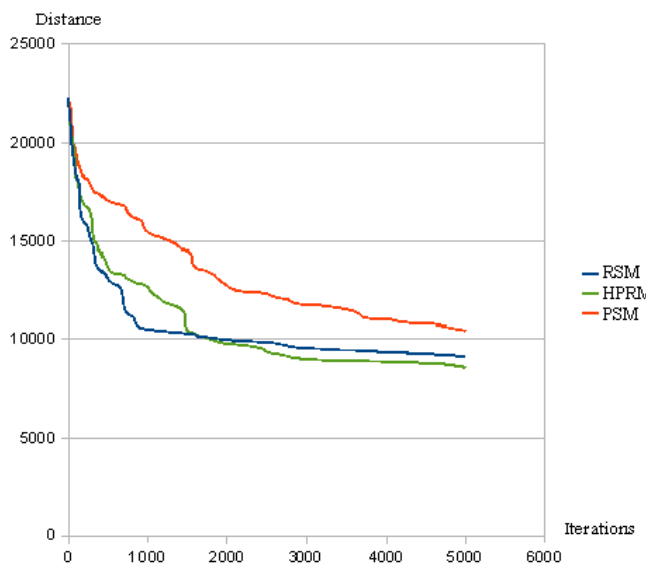

Fig. 7. Convergence comparison of the mutations in Berlin52

## 6. Conclusion

In this paper a new mutation operators known as HPRM (Hybridizing PSM and RSM) is presented. The HPRM operator is a combination of two mutation operators, Partial Shuffle Mutation and Reverse Sequence Mutation The performance of these mutation operators is compared with four existing mutations. For this comparison berlin52, the instance of TSPLIB, is used. The obtained results indicate the performance of the HPRM operator to found the better minimal values than the remaining mutations used in this work.

Based on this study, it is expected that in future the HPRM mutation operator shows a great potential for future research using in other NP-complete problem.